\documentclass[a4paper, 12pt]{article}

\usepackage{latexsym}
\usepackage{graphicx}
\usepackage{amsfonts,amssymb,amsmath}
\usepackage{hyperref}
\usepackage{xcolor}
\usepackage{multirow}
\usepackage{comment}
\usepackage{arxiv}
\usepackage{todonotes}
\usepackage{tikz}
\usetikzlibrary{matrix,shapes,arrows,fit,positioning,calc,backgrounds}
\usepackage{caption}
\usepackage{subcaption}

\newcommand{\RR}{\mathbb{R}}

\def\BibTeX{{\rm B\kern-.05em{\sc i\kern-.025em b}\kern-.08em T\kern-.1667em\lower.7ex\hbox{E}\kern-.125emX}}

\title{Artificial intelligence and renegotiation of commercial lease contracts affected by pandemic-related contingencies from Covid-19. The project A.I.A.Co.}

\author{
Maurizio Parton\\
University of Chieti-Pescara\\
\href{mailto:maurizio.parton@unich.it}{maurizio.parton@unich.it}
\And Marco Angelone\\
University of Chieti-Pescara\\
\href{mailto:marco.angelone@unich.it}{marco.angelone@unich.it}
\And Carlo Metta\\
ISTI-CNR Pisa\\
\href{mailto:carlo.metta@isti.cnr.it}{carlo.metta@isti.cnr.it}
\And Stefania D'Ovidio\\
University of Chieti-Pescara\\
\href{mailto:stefania.dovidio@unich.it}{stefania.dovidio@unich.it}
\And Roberta Massarelli\\
University of Chieti-Pescara\\
\href{mailto:roberta.massarelli@unich.it}{roberta.massarelli@unich.it}
\And Luca Moscardelli\\
University of Chieti-Pescara\\
\href{mailto:luca.moscardelli@unich.it}{luca.moscardelli@unich.it}
\And Gianluca Amato\\
University of Chieti-Pescara\\
\href{mailto:gianluca.amato@unich.it}{gianluca.amato@unich.it}
\And Cristiano De Nobili\\
Pi School\\
\href{mailto:cristiano.denobili@gmail.com}{cristiano.denobili@gmail.com}}

\renewcommand{\today}{}
\begin{document}

\maketitle


\begin{abstract}
This paper aims to investigate the possibility of using Artificial Intelligence (AI) to resolve the legal issues raised by the Covid-19 emergency about the fate of contracts with continuous, repeated or deferred performance, as well as, more generally, to deal with exceptional events and contingencies. We first study whether the Italian legal system allows for `maintenance' remedies to face contingencies and to avoid the termination of the (duration) contracts while ensuring effective protection of the interests of both parties. We then give a complete and technical description of an AI-based predictive framework, aimed at assisting both the Magistrate (during the trial) and the parties themselves (in out-of-court proceedings) in the redetermination of the rent of commercial lease contracts. This framework, called A.I.A.Co. for Artificial Intelligence for contract law Against Covid-19, has been developed under the Italian public grant called \emph{Fondo Integrativo Speciale per la Ricerca}  and - even if the predictive system was initially intended to deal with the very specific problem connected to Covid-19 - the knowledge acquired, the model produced and the research outcomes can be easily transferred to other civil issues (such as, for example, those relating to the determination of the amount of the maintenance or divorce obligation in family law).

\end{abstract}


\section{The A.I.A.Co. project}\label{sec:intro}


This paper aims to investigate the possibility of using Artificial Intelligence (AI) to resolve the legal issues raised by the Covid-19 emergency about the fate of contracts with continuous, repeated, or deferred performance, as well as, more generally, to deal with exceptional events and contingencies. However - even if the predictive system was initially intended to deal with the very specific problem connected to Covid-19 - the knowledge acquired, the model produced and the research outcomes can be easily transferred to other civil issues (see Section~\ref{sec:future_dev}).


\subsection{The Italian legal framework}

The pandemic and the lockdown-type containment measures adopted by the Authority to prevent infections, as extraordinary, unforeseen, and unforeseeable events, can, in fact, be qualified as `contingencies' (\emph{force majeure} and \emph{factum principis}). In particular, jurists had to deal, on the one hand, with the distribution of the contractual risk and, on the other hand, with the management of contingencies from a perspective of preservation of the contract. On this point, see \cite{BenSEI, DolPRT, FedMCP, GroESE, MacDCS, MacSRT}.

The Italian Civil Code provides remedies to face both contingencies and breach of the contract.

As for the first profile, the attention must be first of all directed on the Termination for Supervening Impossibility (art.\ 1256 and 1463 of the civil code and following) and the Termination for Excessive Burden (art.\ 1467 of the civil code). Similar `demolition' remedies are a consequence of the non-fulfillment of the obligation: in this case, the attention must be first focused on the debtor’s liability (art.\ 1218 of the civil code and following) and to the Termination for Breach of Contract pursuant to art.\ 1453 of the civil code and following. These remedies allow only the disadvantaged party to cancel the contractual relationship.

However, the termination of the contract does not always respond effectively to the interests pursued by the parties, who may prefer to continue the contractual relationship, even if under amended provisions. Therefore, it is necessary to verify the presence of ``maintenance” remedies that allow for reaction to contingencies and avoid the termination of the contract, protecting the interests of both parties.

In this regard, Italian legal scholarship has wondered about the configurability of a duty to renegotiate for the parties, even in the lack of an express normative provision – for a review of the different opinions, see \cite{PriACR}. This duty already exists in other systems of civil law, like the German Civil Code (par. 313 BGB about the so-called \emph{Störung der Geschäftsgrundlage}) and the French Civil Code (art.\ 1195 which is about the so-called \emph{imprévision}).

According to part of the Italian jurists, already in the Italian Civil Code, it would be possible to identify solutions aimed at preserving the contract, see \cite{PerGRD, PerDCL, PerCCA, MacARC}; \emph{contra} \cite{GenRSR, SicRin}. In particular, the obligation to renegotiate could be inferred from the general clauses of good faith and fairness pursuant to the art.\ 1175 and 1375 of the Italian Civil Code and from the principles of proportionality, reasonableness, and adequacy, as well as the principle of `social solidarity' pursuant to art.\ 2 of the Italian Constitution, which has been recognized as having direct applicability in relations between private individuals \cite{LipVCP, PerENP, PerPPD}. Thereby, in this perspective, the static and formalist principle of the `\emph{pacta sunt servanda}', codified in the art.\ 1372 of the Italian Civil Code, must be abandoned in favor of the different principle of `\emph{rebus sic stantibus}'. 

In addition, the d.d.l.\ n.\ 1151/2019, as part of the proposed reform of the Italian Civil Code, already provided the right of the parties to obtain the renegotiation of the contract, even before a Court, when the performance of one of the parties has become excessively burdensome \cite{SirEOS}. The Italian Legislator also intervened on this point by introducing during the Covid-19 emergency, even if only in some specific cases, an obligation to renegotiate (see the art.\ 216, paragraph 3, d.l.\ 19/10/2020, n.\ 34, conv.\ l.\ 17/07/2020, n.\ 77 about the reduction of the rental fees in favor of sports facilities for the period of the lockdown; the art.\ 6-\emph{novies} d.l.\ 22/03/2021, n.\ 41, conv.\ l.\ 21/10/2021, n.\ 69, on the right to renegotiate the commercial lease contract for those who have suffered significant reductions in the volume of business during the pandemic period; the art.\ 10 d.l.\ 24/08/2021 n.\ 118, conv.\ l.\ 21/10/2021, n.\ 147, about the negotiated composition of the business crisis).

In the international context, the discipline applicable to the performance of cross-border contracts is oriented towards the affirmation of maintenance remedies with the use of special `hardship' clauses which commit the parties to renegotiate the contract in the event of contingencies. This is also recognized within the \emph{Unidroit} Principles (art.\ 6.2.3), the Vienna Convention of 1980 (art.\ 79), the Principles of European Contract Law (art.\ 6:111), the Common Frame of reference (art.\ 108-110), the European Code of Contracts (art.\ 157) and the Draft United Nations Code of Conduct on Transnational Corporations (art.\ 11). In addition, in 2003 the International Chamber of Commerce elaborated specific `\emph{force majeure}' and `hardship' clauses, updated in 2020 because of the pandemic. Consistently, the Principles for the Covid-19 crisis (n.\ 13), developed by the European Law Institute during the health crisis, suggest that States ensure renegotiation between the parties, even in the absence of a specific contractual clause or a specific legislative provision, pursuant to the principle of good faith.


\subsection{Methods and results}
\par{\textbf{What and why.}} The A.I.A.Co.\ (Artificial Intelligence for contract law Against Covid-19) project started from a competitive research grant – called `\emph{Fondo Integrativo Speciale per la Ricerca}' - FISR 2020 (Project Code: FISR2020IP 04568; CUP: D25F21000500007) – provided by the Italian Ministry of University to resolve the legal issues raised by the Covid-19 emergency. On April 30, 2021, A.I.A.Co.\ passed an initial selection phase, and this led to the creation of the prototype described in this paper. As of December 10, 2022, the prototype is still under evaluation by anonymous reviewers appointed by the Ministry of University for admission to the second part of the funding.

A.I.A.Co.\ final aim is to create a system of `equitable algorithms' for commercial lease contracts, that is, an AI-based framework that can be used to facilitate both judicial or extrajudicial redetermination of the rents and to guarantee the conservation of the contracts, whose balance has been deeply affected by the health emergency.

We restrict to commercial (real estate) lease contracts because this is one of the main topics of the jurisprudential debate that emerged in the aftermath of the pandemic in Italy \cite{CarCLC, PisPLC, DolLEC}. By limiting several business activities, the lockdown-type containment measures placed tenants in difficulty in paying the rental fees. Owners reacted by enforcing the guarantees signed to cover the payments, asking the termination for breach of contract, or acting for the eviction procedure for rent-paying delays. Judges, balancing the opposed interests, in many cases have considered to equitably reduce the amount of the rental fee, attributing the case, alternatively, to the supervening (partial or temporary) impossibility to perform by the owner \footnote{On which see, \emph{ex multis}, Court of Roma, 29/05/2020; Court of Roma, 25/07/2020; Court of Venezia, 28/07/2020; Court of Venezia, 30/09/2020; Court of Milano, 28/06/2021, n.\ 4651; Court of Reggio Calabria, 29/03/2022, n.\ 373; \emph{contra} Court of Roma, 09/09/2020)} or to the obligation to renegotiate the contractual services, inferred from the principles of good faith and fairness in the contract execution phase \footnote{On which, see, \emph{inter alia}, Court of Milano, 08/04/2020, n.\ 2319; Court of Roma, 27/08/2020; Court of Milano, 21/10/2020; Court of Auditors, Emilia-Romagna, 03/06/2021; Court of Palermo, 09/06/2021; Court of Lecce, 24/06/2021; \emph{contra ex multis} Court of Roma, 21/05/2021, n.\ 9457 and Court of Roma, 06/08/2021.}.

\par{\textbf{Who and how.}} The research team was divided into a Law Unit and an AI Unit, which have constantly worked synergically, sharing the results gradually achieved. This organizational structure has proved to be extremely effective in view of project management, especially in consideration of the analyzed issues, which - because of their intrinsic interdisciplinarity - require a combination and integration of the different skills and expertise involved.

Ideally, the A.I.A.Co.\ framework should take the defense deeds as input, and output a decision to be proposed to the Judge, together with explanations of the proposed decision. The decision could be a numerical value, for quantitative legal prediction problems, or a (binary or multiclass) decision. In any case, such a framework should start with a preprocessing phase of the defense deeds, followed by a natural language processing (NLP) phase extracting relevant features, followed by a predictive regression or classification model.
One could wonder why such a complicated design, instead of using an easily implemented predictive model using a spreadsheet. However, such a model would not be learned: it would require experts to choose (\emph{ex ante}) all the different parameters and scenarios, and the underlying rules.

\begin{figure}
\centering
\begin{tikzpicture} 
\node {\includegraphics[scale=0.3]{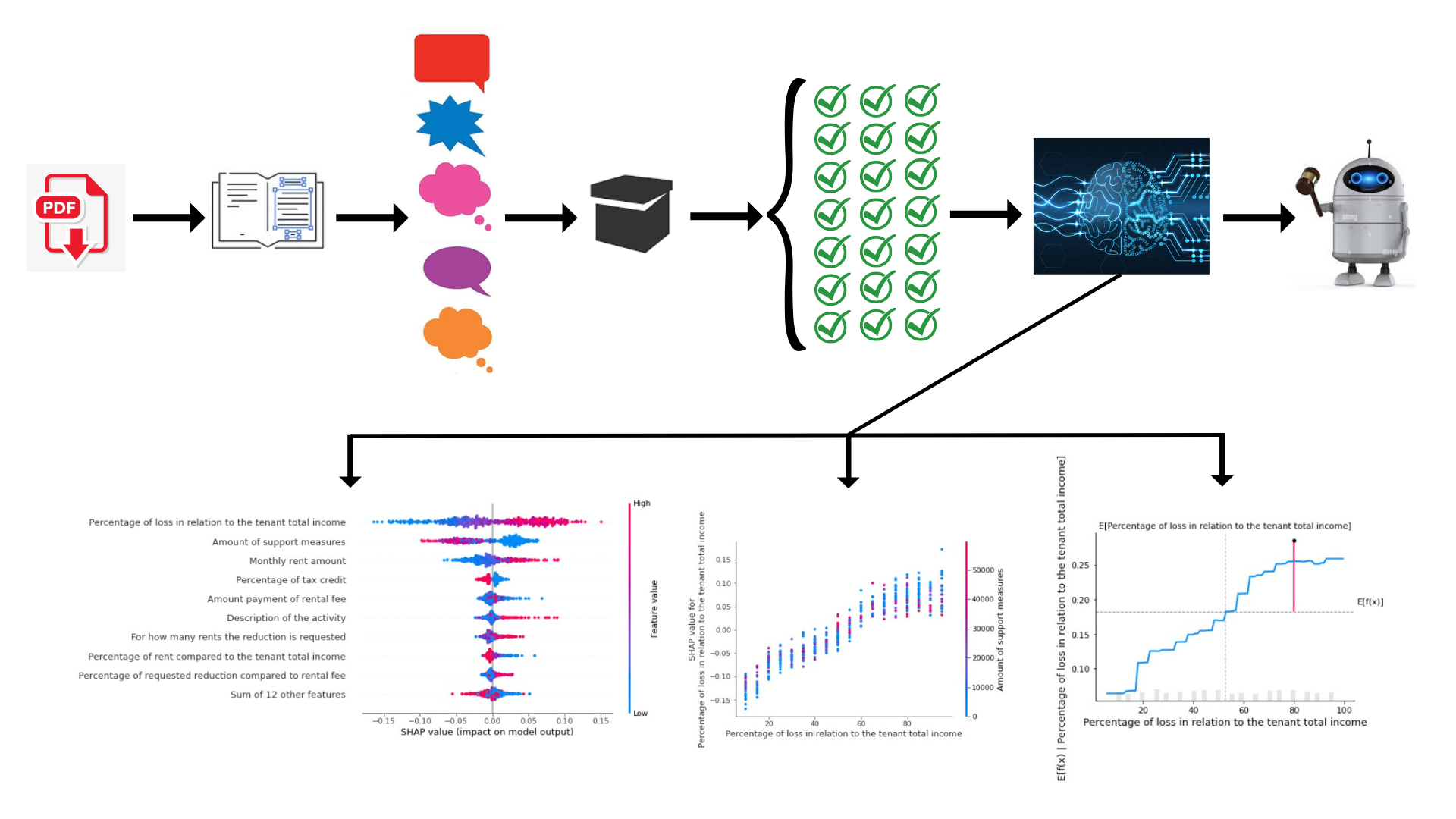}};
\draw [fill=gray, opacity=0.3, rounded corners] (-7.5,0.2) rectangle (0.4,4);
\node at (-5.9,1.7) {1};
\node at (-3.8,1.7) {2};
\node at (-2,1.7) {3};
\node at (-0.1,1.7) {4};
\node at (2.7,1.7) {5};
\node at (5.5,1.7) {6};
\node at (2.8,0.4) {7};
\end{tikzpicture}
\caption{The complete A.i.A.Co.\ framework. In 1, a document containing the defense deeds is preprocessed, producing a text (2) that is then passed to a NLP model (3). In 4, the NLP model outputs the relevant features, which are then passed (5) to the predictive model, which outputs (6) the decision that is then proposed to the Judge. Weights learned during the training of the predictive model are passed (7) to SHAP to produce explanations helping the Judge in using the proposed decision.}
\label{fig:ideal_framework}
\end{figure}
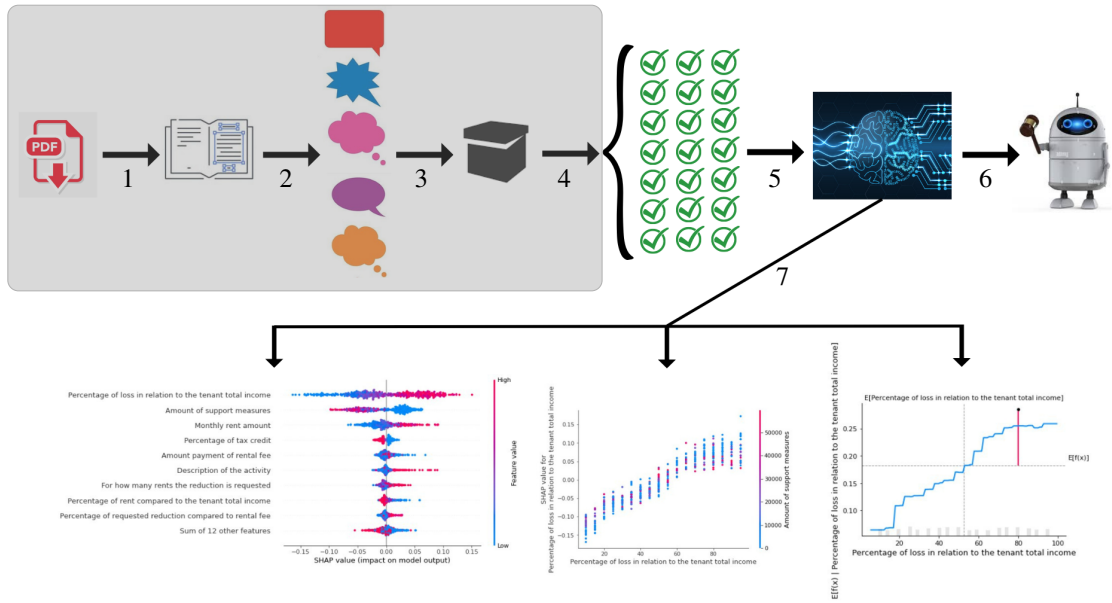

The predictive model should then be analyzed by explainability techniques producing explanations. See Figure~\ref{fig:ideal_framework} for details.

However, the first design decision taken by the two research Units was to concentrate at first only on the final prediction, because NLP is notoriously very expensive in terms of time and resources when fully trained from scratch. We then decided to postpone the implementation of a NLP phase and to adopt, for the prediction phase, training data artificially generated and submitted, for labeling, to authentic Judges.
We point out that, since in this prototype we are interested in the redetermination of the rents, which are numerical quantities, we need to frame the problem as a regression task.

For explanations, we used SHAP. In Figure~\ref{fig:ideal_framework}, the prototype described in this paper is the non-greyed-out part of the picture. In the next paragraphs, we describe in more detail how the training data was produced, the chosen predictive model, and the explainability part.

\par
\begin{figure}
	\centering
	\includegraphics[width=\columnwidth]{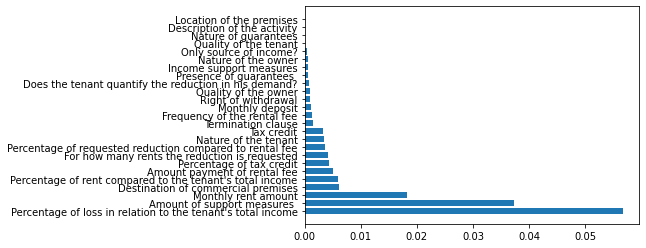}
	\caption{The original 25 expert-selected features, and their corresponding SHAP values. We removed features whose SHAP value was less than $10^{-5}$, and this leaves 21 features.}
	\label{fig:bar_plot_all}
\end{figure}

{\textbf{Generation of training data.}} We started with an expert-based selection of the most informative features that the Judge usually has to evaluate in his decision, identified in 25 fields, see Figure~\ref{fig:bar_plot_all}. In particular, the Law Unit\footnote{This Unit is composed of scholars and PhD students in legal disciplines.}, according to a small group of Judges, selected features through the study and the analysis of the decisions already published. In addition, in the final survey (Section \ref{sec:survey}) we asked Judges, lawyers, scholars, and students in legal disciplines to confirm that elements \footnote{Below one of the artificially generated cases. The parts in italics identify the selected features, that randomly differentiate each case.
`Case

WHEREAS: 

- Mr. Marco Rossi, \emph{a private subject}, is the tenant of a commercial premises in Bari, in which he carries out the activity of \emph{Management of multipurpose sports facilities} (ATECO code \emph{93.1}), \emph{the only source of his income};

- the lessor is \emph{Ente Alfa}, a \emph{private subject}; 

- the contract provides for a monthly rental fee of \emph{5.600,00 euros}, to be paid in \emph{monthly} installments of \emph{5.600,00 euros} each; 

- that rental fee is the \emph{65\%} of the lessor's total annual income; 

- the contract \emph{provides the tenant's right to withdraw} (in addition to the prevision of the Article 27, Paragraph 8, Law No./ 392 of 1978), and \emph{does not provide for an express termination clause, in case of not-fulfilled payment of rent}; 

- the contract, for the fulfillment of the obligations of the tenant, provides for \emph{a guarantee given by a non-professional party}; 

- during the lockdown from Covid-19, \emph{the tenant's income was reduced in 9 months by 85\%};

- the tenant \emph{did not benefit from any income support measures};

- the tenant has obtained - for the period indicated in the law - \emph{a tax credit of 60\% of the paid rental fees}; 

- in the absence of an agreement between the parties on the renegotiation of the contract, the tenant requests to this Judicial Authority the reduction, according to equity, of the amount of the \emph{monthly rental fee for 9 months}. 

FOR THESE REASONS

The Judge [...]'. 

At that point there was a `drop-down menu' with the following alternatives: `\emph{DOES NOT ORDER the reduction of the monthly rental fee}'; or `\emph{ORDERS the reduction of the [...] (with the possibility of indicating a percentage between 5\% and 100\%) of the monthly rental fee'.}}. 
The Law Unit selected a set of realistic constraints to generate likely cases, such as, for example: `monthly rental fee amount' greater than € 500 and less than € 50,000, or: if `nature of the owner' = `natural person' then `quality of the owner' = `private'. Finally, we generated 600 data points, that is, 600 points in $\RR^{25}$, sampled from features distributions with weights chosen again by the Law Unit. Since the Law Unit has strong domain knowledge of the problem we are trying to solve, we expect this human-in-the-loop approach to massively reduce the inevitable distribution shift from the training data to the distribution of actual cases seen by Judges.

Each of these data points was used to produce an artificially generated likely case similar to a defense deed in a human-readable PDF document. We then followed again a human-in-the-loop approach by sending these documents to a selected panel of 30 Judges\footnote{The names of the Judges who consented to the mention are: Giovanna Bilò (Judge at the Court of Ancona, currently placed out of office at the Italian division of the European Court of Human Rights - Strasbourg), Anna Francesca Capone (Judge at the Court of Lecce), Luigi D'Orazio (Judge at the Court of Cassation), Giorgio Di Benedetto (President Emeritus at the Court of Sulmona), Francesca Di Donato (MOT at the Court of Napoli), Alessandro Di Tano (Judge at the Court of Ancona), Salvatore Grillo (Judge at the Court of Appeal of Bari), Rachele Dumella De Rosa (MOT at the Court of Napoli), Mario Fucito (Judge at the Court of Napoli), Martina Fusco (MOT at the Court of Napoli), D.ssa Emanuela Gallo (MOT at the Court of Napoli), Antonio La Catena (Judge at the Court of Foggia), D.ssa Annagrazia Lenti (Judge at the Court of Taranto), Claudia Malafronte (MOT at the Court of Napoli), Dr.\ Guglielmo Manera (Judge at the Court of Napoli), Aldo Marcaccio (MOT at the Court of Napoli), Chiara Maria Marcaccio (MOT at the Court of Napoli), Carlo Picuno (Judge at the Court of Audit), Eleonora Maria Velia Porcelli (Judge at the Court of Milano), Pasquale Angelo Spina (Judge at the Court of Castrovillari), Vincenzo Trinchillo (Judge at the Court of Napoli), Raffaele Viglione (Judge at the Court of Taranto).} for labelling with a percentage representing the amount of the reduction. We opted for the exclusive involvement of professional Judges, almost all belonging to the ordinary jurisdiction ($96.67\%$), to obtain more accurate labels. At the end, we received 557 answers. Subsequently, we use this labeled dataset for training and testing.

\par{\textbf{Models, hyperparameters and training.}} The aforementioned dataset of 557 labeled data points was used to train and test several baselines: a `constant' model predicting always the same percentage of reduction, a model predicting the median reduction on the training set, a decision tree, and a linear model.

We then trained a random forest, with 10-fold cross-validation. This initial training was used to identify and dismiss the less important features. For that purpose, we used a threshold of $10^{-5}$ for SHAP values, resulting in a new dataset with 21 features instead of 25. See Figure~\ref{fig:bar_plot_all} and the explainability paragraph in this section.

\begin{figure}
\centering
\begin{subfigure}[t]{0.45\textwidth}
\includegraphics[scale=0.75]{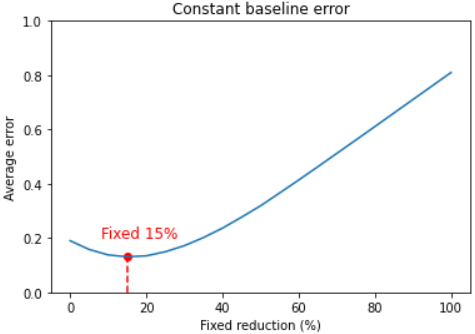}
\end{subfigure}
\hfill
\begin{subfigure}[t]{0.45\textwidth}
\includegraphics[scale=0.75]{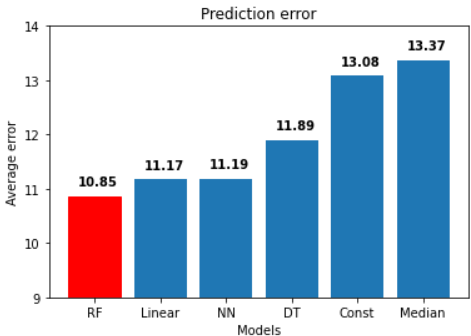}
\end{subfigure}
\caption{Left: mean absolute errors of constant models. The best performance is achieved with a constant reduction of $15\%$.
Right: histogram of mean absolute errors. From left to right: a random forest (RF), a linear model (Linear), a neural network (NN), a decision tree (DT), the best constant model (Const), and a model predicting the median value (Median). In this use case, with synthetic defense deeds and few data points, the performances of RF, Linear, and NN are similar.}
\label{fig:errors}
\end{figure}

Finally, we trained and tested a random forest and a neural network, again with a 10-fold cross-validation as train-test splitting. In predicting the amount of reduction of the rent, the two models turned out to have a similar performance: their absolute mean errors are $0.1085$ and $0.1119$ for the random forest and the neural network, respectively. See Figure~\ref{fig:errors}.

The neural network is fully connected with 21 nodes of input, followed by 3 hidden ReLu layers of 256, 128, and 64 nodes, respectively, and one sigmoidal node for output, for a total amount of 46,849 parameters. The loss is the mean absolute error, trained by the Adam optimizer for 200 epochs and a minibatch size of 32. All other hyperparameters are default Keras values.

The random forest is a \verb|sklearn.ensemble.RandomForestRegressor| with 100 estimators, a minimum samples split of 10, and cross-validation score with scoring = \verb|neg_mean_absolute_error|. All the other default values can be seen at \url{https://scikit-learn.org/stable/modules/generated/sklearn.ensemble.RandomForestRegressor.html}.

The performances of the random forest, the linear model and the neural network are similar. For the explainability part described in the next paragraph, we used only the random forest. With respect to a neural network, a random forest is more easily optimized, more stable, and less opaque. Moreover, all the standard advantages that a neural network could have against a random forest disappear when using so few data points for training. This could possibly change when the feature extraction with NLP will be implemented, see Section~\ref{sec:future_dev}.

\par
\begin{figure}
    \centering
    \includegraphics[width=\columnwidth]{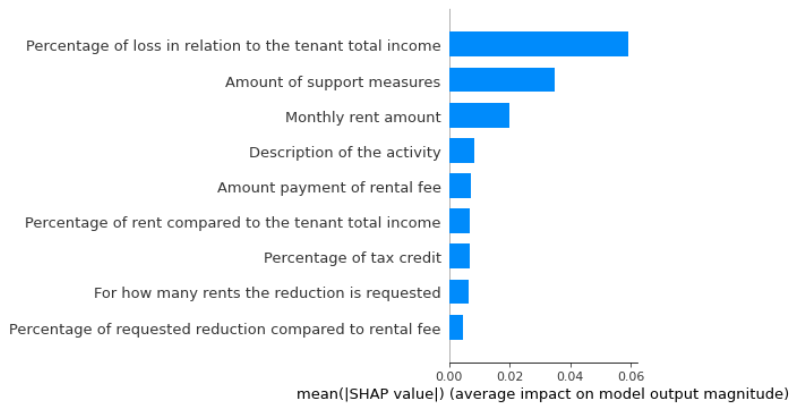}
    \caption{The expected baseline value is $0.17947251$, in figure the top 9 contribution.}
    \label{fig:bar_plot}
\end{figure}

{\textbf{Explainability.}} We release all the code, models, and weights as open source. However, this does not mean that one can understand, by simply reading the code or looking at weights, why a certain decision has been proposed to the Judge. This is because the models we used - random forests and neural networks - are intrinsically black box. Not being able to understand a decision is simply not acceptable if we want a predictive justice framework that can be applied to real-life situations. To overcome this limitation, we used SHAP \cite{LuLUAI}, a state-of-the-art explainability technique that produces explanations, that is, arguments helping human users to understand the proposed decision. In particular, we measure the relative importance of each of the 25 features. In this way, we identify the factual elements that most affected the response provided by the AI. Not surprisingly, the features that turned out to be more relevant to the model decision are `Percentage of loss of income to the tenant's total income', `amount of support measures', and `monthly rent amount', that alone contribute more than $71\%$ to the final prediction, see Figures~\ref{fig:bar_plot_all} and \ref{fig:bar_plot}. Several other SHAP-derived explanations can be found in Section~\ref{sec:explainability}. These explanations can be used in real life to overcome objections regarding the opacity of AI.

We remark that we could have used a simple and interpretable by design model (a linear model or a decision tree), as shown by the baselines in Figure~\ref{fig:errors}. However, we plan to use A.I.A.Co.\ to deal with much more complex use cases, that will very unlikely be described by a linear model.

A prototype of the software described in this paper can be tested at \url{https://vlab3.unich.it/aiaco}, in Italian. All the source code, models, and training data will be available at \url{https://github.com/MistyDay86/A.I.A.Co}.

\section{Related work}

How to use machine learning to estimate Judges' decisions? This question is crucial in predictive justice, and can now be efficiently tackled thanks to the growing availability of legal big data. In the following, we list a few papers that in our opinion share some ground with our project. This list is by no means complete.

In \cite{ChECML}, half a million asylum hearings rendered by 441 unique judges over a 32-year period, from 336 different hearing locations, are analyzed. The authors consider the binary task of granting / not granting asylum and use a random forest for classification. In \cite{ATPPJD}, NLP is used to predict the binary outcome (violation yes/no) of cases tried by the European Court of Human Rights based solely on textual content. \cite{MWVRFA} is a review of NLP methods applied to classification tasks in legal prediction.
As in \cite{ChECML}, we use a random forest in the prototype, and we will use NLP as in \cite{ATPPJD, MWVRFA} in the next version of the software. However, we frame the problem as a regression task, instead of classification, because we need to predict a numerical quantity in $[0,1]$ (the reduction rate).

In the seminal paper \cite{KorPSC}, the author develops a formula to extract numerical features that are then used to determine a decision boundary. This is not machine learning, but as far as we know is one of the first contributions to the field of predictive justice, and in some sense is concerned, like us, with numerical quantities and not only classification.

Then, Daniel Martin Katz in \cite{KatQLP} coined the term `quantitative legal prediction', trying to respond to questions like `Do I have a case? What is our likely exposure? How much is this going to cost? What will happen if we leave this particular provision out of this contract? How can we best staff this particular legal matter?'. Our research goes in this direction by trying to answer the question: `how much is the reduction of the rental fee?'.
As a possible future development, when the NLP phase will be completed, we plan to extend the A.I.A.Co.\ framework to other quantitative legal questions, like, for example, `how much is the maintenance allowance?'. See Section~\ref{sec:future_dev} for more details.


A recent interesting paper approaching quantitative legal prediction is \cite{CAANLP}. The authors use attention-based NLP models to predict the importance of a case on a scale from 1 to 4 in a regression task. To train and evaluate their models, the authors develop and release a very useful English legal judgment prediction dataset \cite{CAADat}. Similar to what we plan to do in the final version of our project, they use NLP; however, A.I.A.Co.\ is an end-to-end framework that will process legal documents starting from their original format (e.g.\ PDF), extracting both the input features and the label to be used during training. In this sense, A.I.A.Co.\ is an instance of self-supervised learning \cite{LeMMet}.

\section{Validation survey}\label{sec:survey}

We designed a survey with the aim of evaluating the potential impact on the justice system given by our prototype model, as well as the effectiveness of explanation methods as a tool to assist Judges.

\subsection{Survey structure}

The survey is organized into five question panels, from \textbf{Q1} to \textbf{Q5} in the following.
\\
\textbf{Q1}. The first panel is used to collect general personal data, their profession and degree of experience and request informed consent from participants.
\\
\textbf{Q2}. In the second panel, the participants are presented with a synthetic defense deed and the corresponding decision predicted by the prototype model. Participants are asked if they are satisfied with the presentation of the case, with the quantity and quality of the information contained in the premises, and with their plausibility. Finally, to what extent do they agree with the decision predicted by the model.
\\
\textbf{Q3}. The third panel is devoted to the presentation of the three most informative features according to the SHAP model. Participants are asked to what extent they agree that the features presented are the most relevant to the case under discussion, on the values assigned by SHAP, on the confidence towards the predicted decision after having seen the SHAP explanation, and, finally, how useful they find having the additional information about feature importance available in a real scenario.
\\
\textbf{Q4}. In the fourth panel, we present three different counterfactual instances. Each instance is derived from the original one by selecting the 3 most informative features presented in the previous panel as a set of actionable features, one at a time for each counterfactual explanation. For this, we use the BF counterfactual method in the \emph{Fat Forensic Library}\footnote{\url{https://fat-forensics.org/}}. Participants are then asked to rate their confidence towards such counterfactual instances and, finally, how much they consider useful to have a model available from which to request counterfactuals such as those just described.
\\
\textbf{Q5}. In the final panel, we ask for overall assessment and confidence in predictive justice models as seen in the case under discussion, leaving a blank field at the end of the survey to write comments, suggestions, and criticisms related to the use case.
\\
In all quantitative questions, participants can choose their rate on a scale of 1 to 5.

\subsection{Hypothesis}

The structure of the investigation reflects the following
hypothesis we intend to address.
\begin{itemize}
\item \textbf{H1}. The synthetic defense deed is satisfactorily presented in its contents, in the order and structure of the information.
\item \textbf{H2}. Participants generally agree with the predicted decision and, moreover, show greater confidence in that decision after receiving the explanations.
\item \textbf{H3}. Participants find counterfactual instances a useful aid tool for assisting in real scenarios.
\end{itemize}

\subsection{Survey results}

\begin{figure}[]
  \centering
  \subfloat{\includegraphics[width=0.4\textwidth]{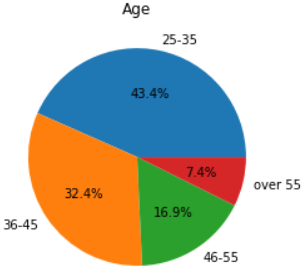}\label{fig:f1}}
  \hfill
  \subfloat{\includegraphics[width=0.44\textwidth]{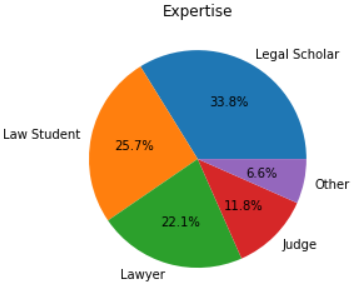}\label{fig:f2}}
  \caption{Aggregate demographic statistics of the participants.}
  \label{fig:anagraphic}
\end{figure}

A total of 138 participants took part in the study. Participants signed up for the survey online after digitally signing a consent form followed by a short demographic survey. Aggregate demographic statistics of the participants are available in Figure \ref{fig:anagraphic}. All participants have a legal background and, among them, as many as 67.7\% belong to key categories for our survey, such as Judges, legal scholars, and lawyers.

\begin{figure}[!tpb]
  \centering
  \subfloat[]{\includegraphics[width=0.5\textwidth]{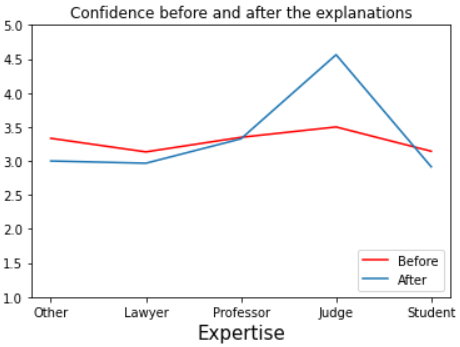}\label{fig:ff1}}
  \hfill
  \subfloat[]{\includegraphics[width=0.5\textwidth]{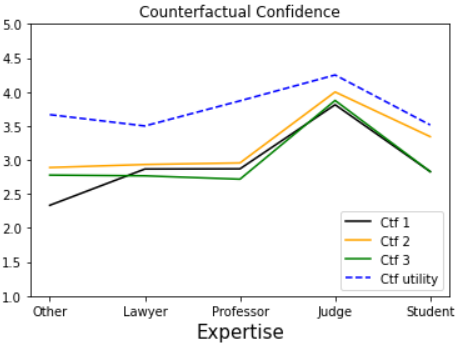}\label{fig:ff2}}
  \caption{Participants' confidence in (a) the model prediction before and after looking at the SHAP explanation and in (b) the counterfactual instances.}
  \label{fig:confidence_before_after}
\end{figure}

Figure \ref{fig:ff1} shows the participants’ confidence in the model prediction before and after looking at the explanations returned by SHAP. While there is no appreciable impact in receiving explanations from some categories, a marked increase in confidence is noted instead if the analysis is restricted to the judges, the most significant category. However, we observe that the confidence reaches at least the rating of $3.0$ for all categories of participants. This confirms to some extent the hypothesis \textbf{H2}, more strongly if restricted to the most meaningful category.
Figure \ref{fig:ff2} shows the participants’ confidence in counterfactual instances. As before, there is a clearly better reaction from the judges, while the other categories still show a good rating level. Moreover, participants rate the usefulness of having counterfactuals available with an average score of $3.72$. The such score is even more promising if restricted to judges ($4.25$), while it takes its lowest value of $3.50$ among law students. This analysis addresses hypothesis \textbf{H3}.
Hypothesis \textbf{H1} is instead addressed by observing the answers to the first part of question \textbf{Q2}. Participants score an average value of $3.62$, while judges' scores climb up to $4.37$.
All the described analyses also show unequivocally that the judges achieve the best score in all the questions that were addressed to the participants.

\section{Impact to justice}
\label{sec:predictive_justice}

\subsection{Discretionality}
The framework developed in this project gives the Judge a reference tool to estimate the rental fee reduction in favor of tenants. The software/prototype described in the paper does not aim to predict all the \emph{force mejer} situations: it intends to offer an equitable solution for renegotiation in case of contingencies (whatever they may be).

This mechanism could guarantee greater uniformity of the judgment and legal certainty. A similar system is already used in Italy to determine the amount of compensation for non-pecuniary damages, see `\emph{Tabelle di Milano}', elaborated by the Observatory of the Court of Milano and used throughout the Italian national territory, available at \url{https://www.tribunale.milano.it/files/Tabelle%20milanesi_Danno%20non%20patrimoniale_ed.%202021.pdf}.

The aforementioned tables, however, do not eliminate the discretion of the Judge, who can always `customize' the decision proposed by the algorithm, adequately justifying his decision.

Indeed, also in our case, the Judge, starting from the A.I.A.Co.\ suggestion, must compensate for any bias or other possible limitations of the software in a corrective way. This is also in consideration of the fact that the decision must in any case be attributable to the Judge, who is ultimately responsible.

Moreover, the law is a lens on reality: the Judges are called to interpret the law in the light of historical, sociological, and political changes at the moment of the decision. Otherwise, the machine is unable to grasp these aspects, because the `predictive justice' originates from the judicial decisions already published. Thus, it appears evident the indispensability of the Judge-natural person, who is called to re-read the decision proposed by the algorithm. Generally, on the issues raised by predictive justice on the Italian scene, see \cite{DalCCP, GabCNG, MatDNG, RulGPI, VioILM}.

Therefore, the AI suggestion must be only a `second opinion', which helps to legitimize the introduction of the framework despite its possible weaknesses. Thus, also, the dominant opinion, see \cite{ACFAPJ}.

On the other hand, human decision is also affected by bias. As Holmes points out, legal certainty is never given because it depends on the foreseeability of human action and human interpretation (that is, in the civil proceedings, the Judge), both of which are fundamentally uncertain. On this point, see also \cite{HilLCE, KBBGAP, KurSNW, AlaPLT}.

It would be necessary/useful because, currently, the Italian Judges should decide the reduction of the rental fee (which has become clearly excessive) according to equity. One of the criticisms of this type of solution (i.e. of an equity-based decision) depends on the fact that Judges normally decide without having parameters to guide their decision (our prototype offer a suggestion that takes into account similar rulings decided before by other Judges).

For these reasons, a man-machine integrated system allows us to overcome both the human and the artificial limits, proposing a decision as fair as possible. More generally, on the importance of the man-machine integration see \cite{PasAPU, PasNLR}. Thereby, it is also possible to avoid the application of the rules on fully automated decision-making systems provided by EU Reg. 2016/679 so-called G.D.P.R.

At the present, the use of A.I.A.Co.\ would not be intended as mandatory for Judges. The discretion of the Judge should not be eliminated, as in the case of the above-mentioned `\emph{Tabelle di Milano}'. The Judges could therefore deviate from the proposed solution if they do not share its application assumptions. In fact, the framework would be applicable only if shared by the Judge, and this is always in view of a Judge-centric justice.

\subsection{Accountability} 
Of course, even if the Judge uses the proposed solution, he should still justify the provision and the solution adopted in the specific case: in fact, both the availability of the source code and models, and the explainability mechanism used, do not replace the Judge's duty to justify, but instead allows the parties and the Judge himself to understand how and why a certain decision has been proposed. In particular, the multidisciplinary nature underlying the design of A.I.A.Co. - a design that requires not only legal skills but also technical, mathematical, and statistical ones - does not exempt the Judge from accompanying the technical description of the framework with a motivation for the decision, making it legible and comprehensible to the parties. This allows, as already mentioned, the parties to know the assumption and review the result. In general, on the importance of the collaboration between jurists and computer scientists for technology able to protect human rights, see also \cite{HilLIE}.

\subsection{Transparency and clarity}
The mechanism through which the decision is made must be known to the parties in advance, also considering the principle of transparency and clarity. This knowability must concern its authors, the procedure used for its elaboration, the decision mechanism (including the priorities assigned in the evaluation and decision-making procedure), and the data selected as relevant. On this point, see also \cite{DivDDL, GolPCL}. This is because the parties must be able to verify that the criteria, conditions, and results of the procedure comply with the requirements and purposes established by the law upstream of this procedure. In this way, the parties can also verify if the methods and rules used are clear. Furthermore, regarding the verification of the results and the relative imputability, the downstream verification must be guaranteed, in terms of logic and correctness of the results (on this point, see also Cons.\ State n.\ 8472/2019).

Every component of A.I.A.Co.\ is open source, precisely to allow the parties and the Judge to perform a thorough analysis before deciding whether to use it. This also allows overcoming any objections about the intrinsic opacity of the models used. On the problem of the opacity of AI systems, see also \cite{HilLCE, PasBBS}.

\subsection{Duration of proceedings}
In addition, judicial use of A.I.A.Co.\ in this kind of trial may affect the duration of the proceedings - a problem particularly felt in Italy - speeding up decisions. In fact, the Judge is facilitated in identifying the criteria on which to parameterize the decision and has an instrument that proposes the decision itself.
This is also in accordance with the principle of `fair trial' granted by art.\ 111 of the Italian Constitution and by art.\ 6 ECHR. In particular, the Court of Strasbourg has repeatedly specified that among the guarantees of a fair trial, there is also respect for the reasonable length of the trial, as an instrument to guarantee the efficiency and credibility of justice (see European Court of Human Rights, 24 October 1989, H v.\ France).

On the other hand, in the non-litigation phase, A.I.A.Co.\ acts as a tool to decrease the number of cases litigated before the Courts, with inevitable economic benefits for the parties. This is especially because the parties may predict and share the A.I.A.Co.\ decision before contacting the professional Judge. 

In this way, this predictive framework can, potentially, facilitate access to justice, from a perspective of democratic justice, and ensure the effective protection of human rights; thus, also in accordance with Goal 16 `Peace, justice and strong institutions' of the 2030 Agenda for Sustainable Development, signed in September 2015 by the governments of the UN member countries.

\subsection{Appeals system}
Regarding the appeals system provided by the Legislator, indeed, the algorithm does not change the current discipline. For judicial measures that base their decision on the A.I.A.Co.\ prediction, the ordinary appeal system would remain valid. As regards, instead, A.I.A.Co.\ used as an out-of-court remedy, if the parties do not share the proposed solution, an agreement will not be reached and the parties will have to contact the professional Judge, as already happens in current systems by Alternative Dispute Resolution.

\section{Limitations and future developments}
\label{sec:future_dev}

One of the future research directions that originate from this Project is the promotion of Artificial Legal Intelligence at the service of the rule of law. In this way, as already stated by Hildebrandt, we could pass from `legal by design' to `legal protection by design' \cite{HilLCE}.


The subjects involved in the Project have already highlighted the potential of the device. In particular, almost all the Magistrates contacted have expressed interest in experimenting and using the final prototype result in their own Judicial Offices.

Clearly, the fact that as of now the A.I.A.Co.\ project is missing the initial NLP feature extraction part, is its most severe limitation. However, this prototype provides essential information for the future development of the NLP model: for example, SHAP suggests that the salient features to be extracted are likely to be less than 10.

In the future we will implement the extraction of features with NLP techniques, exploiting the data already available at the Italian Ministry of Justice thanks to the so-called PCT (Telematic Civil Process), appropriately anonymized. From this point of view, this data will grow thanks to the Italian National Recovery and Resilience Plan, so-called N.R.R.P. In particular, the Ministry of Justice, within the `Digitization, Innovation and Security in the P.A.' mission funded by the N.R.R.P., proposes to digitize about 10 million hybrid and paper Court dossier of the proceedings for the years 2016-2026 (on which see https://padigitale2026.gov.it/misure/). Outside the Italian national panorama, systems of quantitative analysis of jurisprudence are already used (e.g.\ U.S.A., Belgium, Czech Republic, France, Germany, and Israel), so this system will be easily exportable abroad. On this point, see \cite{MVWUML}. The ISTI-CNR of Pisa (\url{https://www.isti.cnr.it/en/}) has proposed to collaborate in this future NLP step.

Even if this predictive framework was initially intended to deal with a very specific problem (the lockdown determined by Covid-19), potentially, the knowledge acquired, the model produced and the research outcomes can be easily transferred to other civil issues. The reference is, for example, to the following questions:  family law (e.g.\ to determine the amount of the maintenance or divorce obligation); the calculation of the compensation for damages, including in the insurance field; executive and insolvency procedures (e.g.\ as regards to the determination of allocation plans; moreover the N.R.R.P.\ reforms the negotiated composition for the solution of business crisis, establishing a software capable of verifying the sustainability of debts and automatically drawing up a recovery plan for liabilities below a certain threshold, see art.\ 30 d.l.\ n.\ 118/2021, conv.\ l.\ n.\ 147/2021, and art.\ 30-ter and 30-quinquies of d.l.\ n.\ 152/2021, conv.\ l.\ n.\ 233/2021); renegotiation of business contracts as part of the proceedings of the negotiated composition of the business crisis pursuant to art.\ 10 of d.l.\ n.\ 118/2021, conv.\ l.\ n.\ 147/2021.

Furthermore, the potential of A.I.A.Co.\ increases when we consider the projects active on the international scene that aims to translate the rules into self-executing codes. On which, see, \emph{ex multis}, \cite{MCPCPL, MoRCrC, DivDDL}. In fact, in that case, it will be even easier to use A.I.A.Co.

Finally, an interesting question has been raised by a referee. Since we are dealing with situations that are by nature exceptional, this exceptionality seems antithetical to predictive automated calculations. 

As mentioned above, the software/prototype described in the paper does not aim to predict \emph{force majeure}/exceptional situations: it intends only to offer an equitable solution for renegotiation in case of contingencies. 

We do not exclude the possibility of exploring the contradiction between exceptional situations and predictive automated calculation in future research.

\section{Conclusions}

This predictive framework has wide potential, going so far as to entrust the decision of concrete issues to AI systems, unlike the other similar Italian initiatives. The reference is to research of the University of Brescia, University Sant'Anna of Pisa, and \emph{Alma Mater Studiorum} of Bologna available at \url{https://giustiziapredittiva.unibs.it/home}, \url{https://site.unibo.it/cross-justice/en/project-results/tools} and \url{https://www.predictivejurisprudence.eu/}, respectively. Therefore, our project can represent an important advance for the research about legal prediction and can potentially be used, with minor adaptations, to a wide range of situations, such as those mentioned above.

However, despite the wide potential of A.I.A.Co., the decision – as mentioned in Section~\ref{sec:predictive_justice} – must still be Judge-centric, that is, subject to human control, as required by the anthropocentric approach. This approach inspires also the European strategy on AI, expressed in the White Paper on AI of 19/02/2020, COM/2020/65, and in the Proposal for a Regulation on `harmonized rules on AI', COM/2021/206 (on this point see also the work of the European Commission for the Efficiency of Justice - CEPEJ available at \url{https://www.coe.int/en/web/cepej}). Therefore, the whole research project was inspired by this perspective, without questioning the irreplaceable nature of the Judge-natural person in the use of AI systems, and so also in accordance with the provisions of the Italian National Research Program-P.N.R.\ 2021-2027, which aims at a so-called human-centered innovation.

The NLP version of A.I.A.Co.\ could assist not only the Judge but also his auxiliaries (such as, for example, the professional delegate in executive procedures), the same parties, and the other actors of the out-of-court phase (lawyers, mediators, conciliators, experts, and insurance companies, \emph{etc.}), with a view to deflation of the disputes. Generally, about the fact that technology may assist lawyers, see also \cite{CAANLJ, SteWEU}.

A.I.A.Co. is more flexible than other similar AI tools, overcoming the known issues in the field of predictive justice, without questioning the irreplaceable nature of the Judge-natural person in the use of AI systems.

A.I.A.Co.\ is, therefore, in harmony with the objectives of modernization and rationalization of the judicial system pursued by the recent reform of the Italian civil process (l.\ n.\ 206/2021, so-called `Cartabia reform'), by the establishment of the `\emph{Ufficio per il processo}', whose primary function is precisely that of guarantee the reasonable duration of the process, thanks to the use of technology and to support the innovation processes in the judicial offices, see art.\ 16-\emph{octies}, d.l.\ n.\ 179/2012, conv.\ l.\ n.\ 221/2012 and \url{https://www.giustizia.it/giustizia/it/mg_2_9_2.page}, as well as by the Italian National Recovery and Resilience Plan, which aim to favor the resolution of disputes and the use of technology in order to improve the efficiency of the process and lighten the workload of the Magistrates. The results obtained, then, can represent a valid and immediate tool for use in the Judicial Offices, as well as in other Alternative Dispute Resolution (e.g.\ the Arbitration Chambers or the mediation or conciliation bodies and others operating within the various Councils of the Bar Associations, Chambers of Commerce, \emph{etc.}).

Ultimately, the project testifies that man-machine integration cannot be ignored in order to achieve fair justice. In fact, the use of AI in predictive justice systems must not be demonized or absolutized, thus avoiding embracing the dangerous extremism of techno-optimism and techno-pessimism. In this vein, the prior interconnection between the world of law and that of technology is necessary, which must dialogue and complement each other.

\clearpage

\bibliography{mau.bib}
\bibliographystyle{alpha}

\nocite{AlaPLT, BenSEI, CAANLJ, DalCCP, DolPRT, DolLEC, FedMCP, GabCNG, GenRSR, GroESE, HilLIE, KBBGAP, KurSNW, LipVCP, MacDCS, MacSRT, MatDNG, MVWUML, PasAPU, PasBBS, PasNLR, PerENP, PerPPD, PisPLC, RulGPI, SicRin, SirEOS, SteWEU, VioILM}

%
%


\section{Supplementary material}
\label{sec:explainability}

The interested reader can find in this section a sample of the defense deeds that have been sent to the Judges for labeling, see Section~\ref{sec:intro}, and several figures produced with SHAP, see Section~\ref{sec:explainability}. These figures can be used to understand from various dimensions why the A.I.A.Co.\ framework outputs a certain decision.

\begin{figure}
	\centering
	\includegraphics[width=0.72\columnwidth]{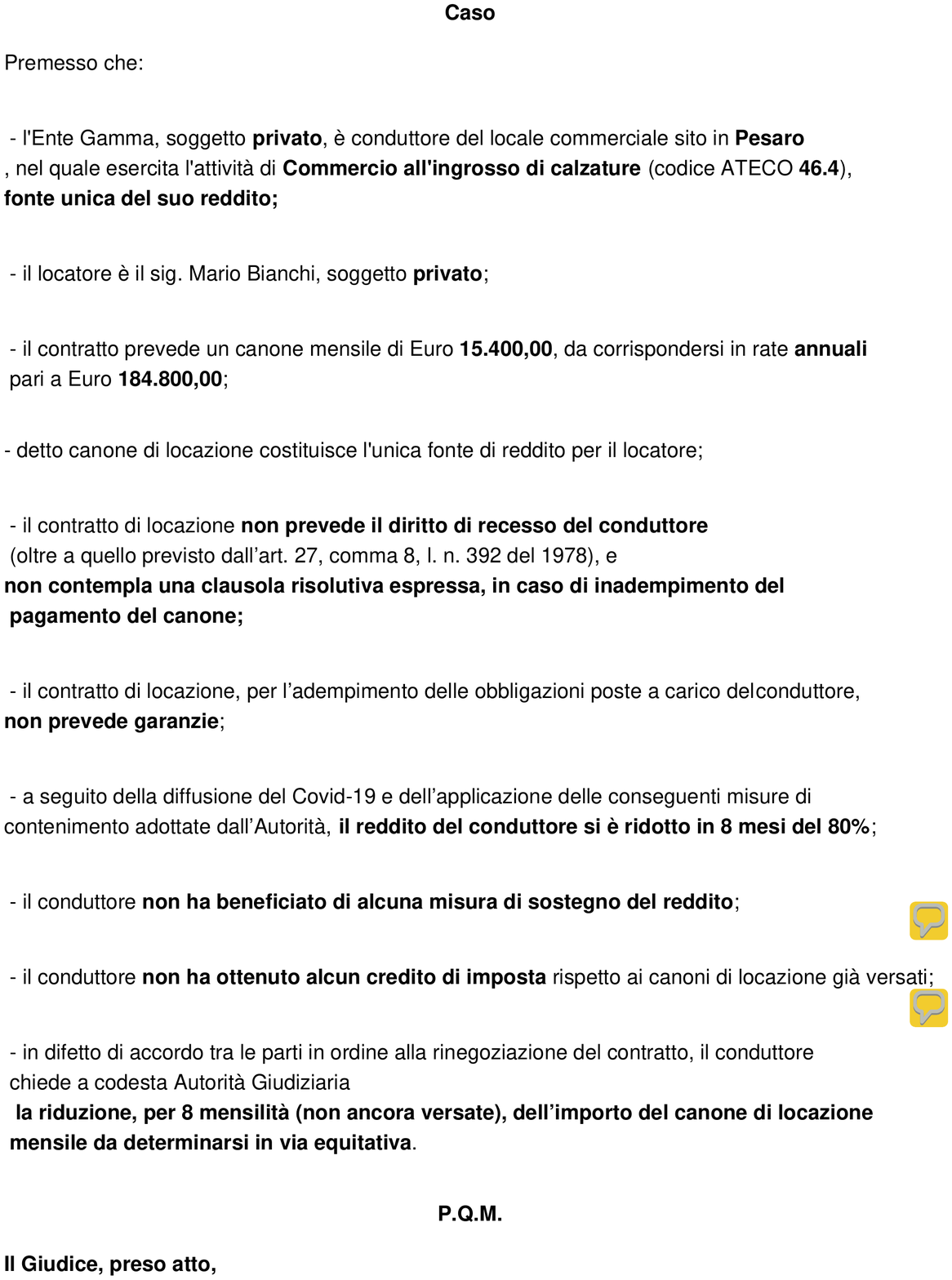}
	\caption{One of the artificial defense deeds sent to the Judges. The Law Unit selected a set of realistic features and likely constraints, together with weights, to attenuate the distribution shift from the training data to actual cases. Then, 600 data points have been sampled using the chosen feature distributions. Finally, each of these data points was used to produce an artificially generated defense deed in a human-readable PDF document in the Italian language. This is the first of these 600 defense deeds, as it was transmitted by email to a Judge.}
	\label{fig:defence_deed}
\end{figure}

\clearpage

\begin{figure}
    \centering
    \includegraphics[width=0.8\columnwidth]{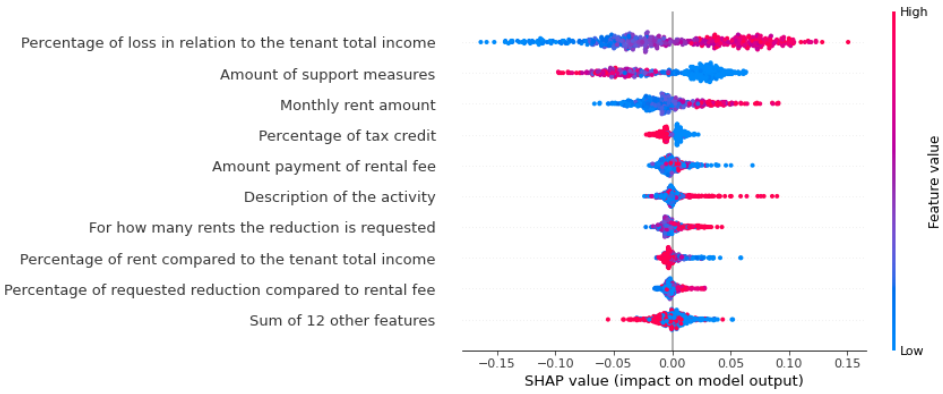}
    \caption{Scattered plot with SHAP values of all instances for the most influential features. The values are grouped by the features on the $y$-axis. For each group, the color of the points is determined by the value of the same feature. The features are ordered by the mean SHAP values.}
    \label{fig:bees_plot}
\end{figure}

\begin{figure}
    \centering
    \includegraphics[width=0.8\columnwidth]{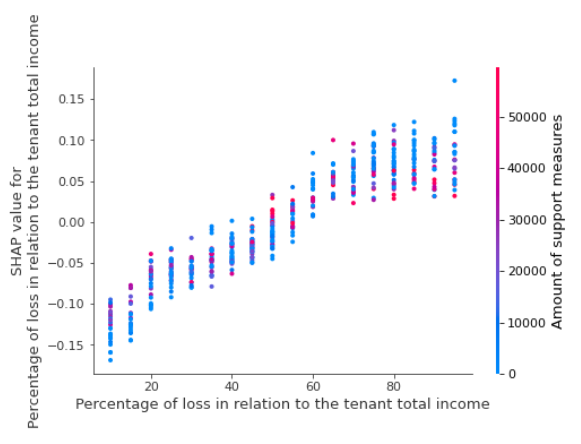}
    \caption{Dependence plot for the shape values of the feature `Percentage of loss in relation to the tenant income', the most important feature.}
    \label{fig:dependence_plot_1}
\end{figure}

\begin{figure}
    \centering
    \includegraphics[width=0.7\columnwidth]{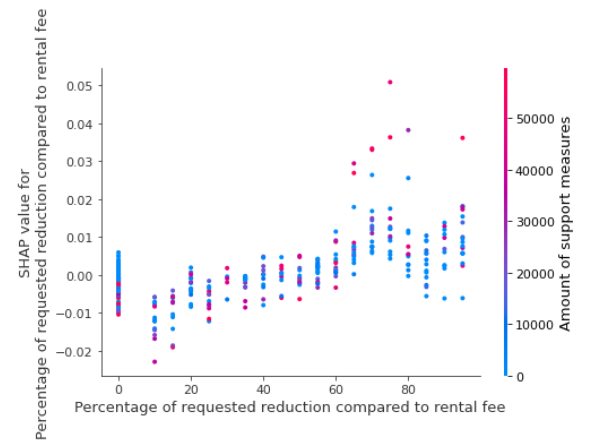}
    \caption{Dependence plot for the shape values of the feature `Percentage of requested reduction compared to rental fee', one of the least important features.}
    \label{fig:dependence_plot_2}
\end{figure}

\begin{figure}
    \centering
    \includegraphics[width=\columnwidth]{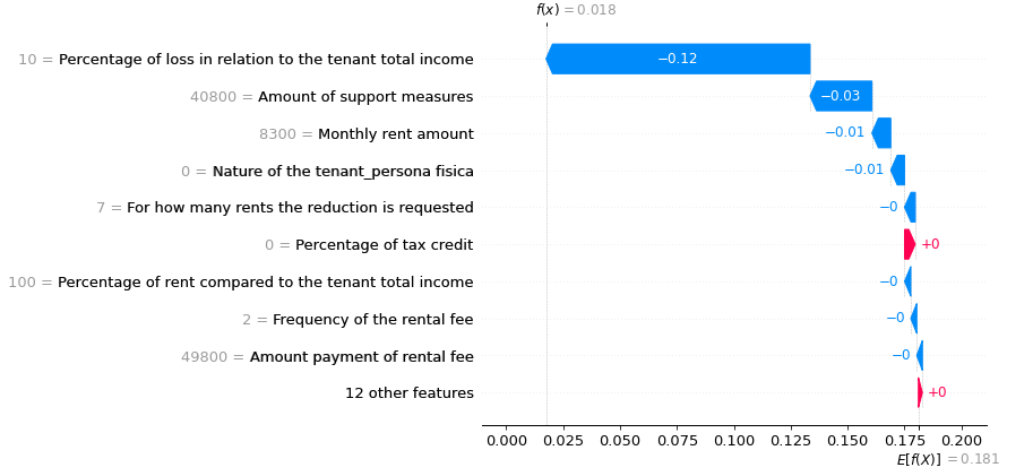}
    \caption{Waterfall plot. Shapley values always sum up to the difference between the game outcome when all features are present and the game outcome whit no features at all. Thus, SHAP values of all the input features will always sum up to the difference between the baseline (expected) model output and the current model output for the prediction being explained. The easiest way to see this is through a waterfall plot that starts at our background prior expectation $\mathbb{E}[f(X)]$, and then adds features until we reach the current model output $[f(X)]$.}
    \label{fig:shap_value_instance_1}
\end{figure}

\begin{figure}
    \centering
    \includegraphics[width=\columnwidth]{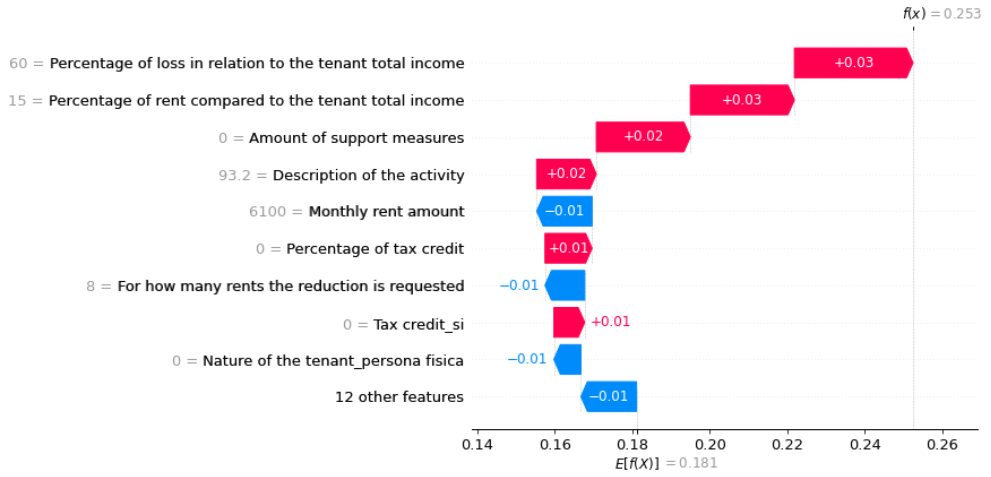}
    \caption{Same as the previous figure, but referred to another instance.}
    \label{fig:shap_value_instance_2}
\end{figure}

\begin{figure}
    \centering
    \includegraphics[width=\columnwidth]{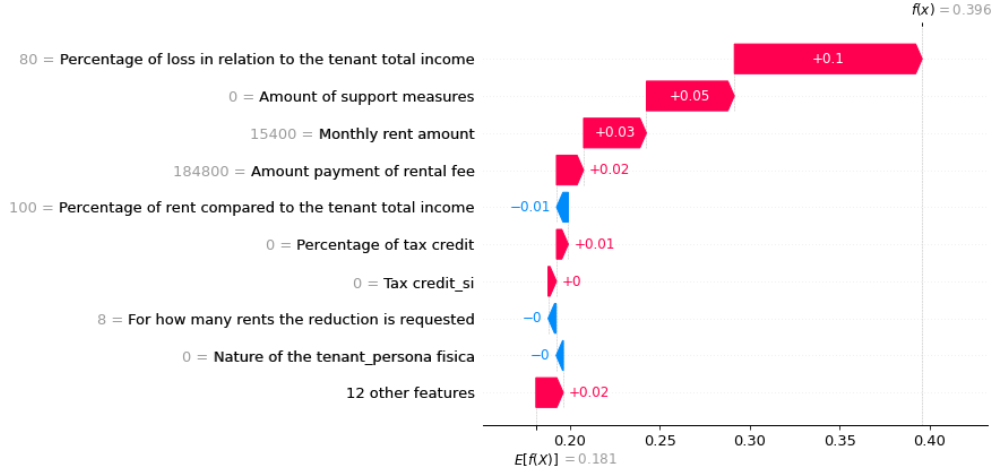}
    \caption{Same as the previous figures, but referred to another instance.}
    \label{fig:shap_value_instance_3}
\end{figure}

\begin{figure}
    \centering
    \includegraphics[width=\columnwidth]{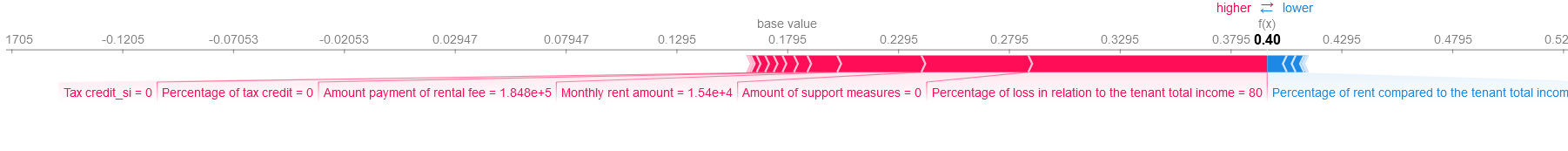}
    \caption{Force plot. These give us the same information as a waterfall plot. We start at the same base value. You can see how each feature increases/decrease the predicted value to give us the final prediction of 0.40}
    \label{fig:straight_plot}
\end{figure}

\begin{figure}
    \centering
    \includegraphics[width=0.9\columnwidth]{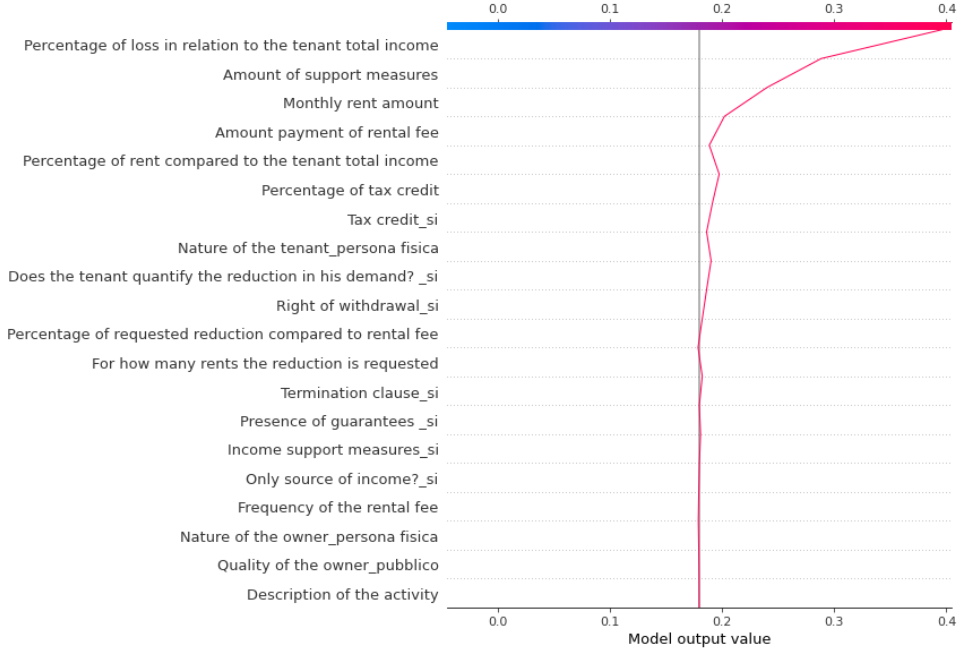}
    \caption{To understand how our model makes predictions we need to aggregate the SHAP values. One way to do this is by using a decision plot. In Figure we can see there is one line in the plot. It starts at the same base value and ends at its final predicted number. As you move up from each feature on the y-axis, the movement on the x-axis is given by the SHAP value for that feature. This gives you similar information to a waterfall plot.}
    \label{fig:tree_plot_single}
\end{figure}

\begin{figure}
    \centering
    \includegraphics[width=0.9\columnwidth]{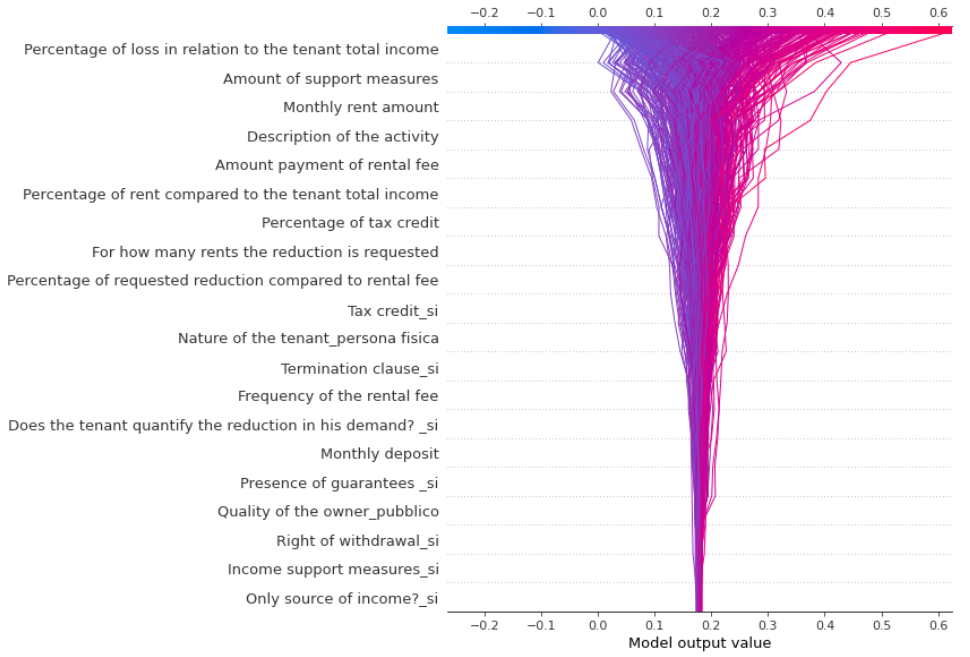}
    \caption{Same as before, but with all the instances at the same time.}
    \label{fig:tree_plot}
\end{figure}

\begin{figure}
    \centering
    \includegraphics[width=0.6\columnwidth]{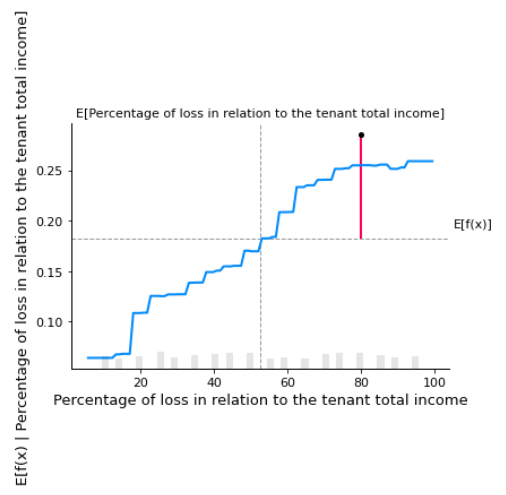}
    \caption{Scatter additive plot for the most important feature.}
    \label{fig:additive_plot}
\end{figure}

\end{document}